\documentclass[11pt,a4paper]{article}
\usepackage[hyperref]{emnlp2020}
\usepackage{times}
\usepackage{latexsym}

\usepackage{microtype}

\usepackage{bm}
\usepackage{amsmath}
\usepackage{amssymb}
\usepackage{graphicx}
\usepackage{url}
\usepackage{multirow}
\usepackage{color}
\aclfinalcopy 
\def\aclpaperid{***} 

\title{Syntactic and Semantic-driven Learning for \\Open Information Extraction}

\author{
\textbf{Jialong Tang}${}^{1,3}$, \textbf{Yaojie Lu}${}^{1,3}$, \textbf{Hongyu Lin}${}^{1}$, \\
\textbf{Xianpei Han}${}^{1,2,*}$, \textbf{Le Sun}${}^{1,2,*}$, \textbf{Xinyan Xiao}${}^{4}$, \textbf{Hua Wu}${}^{4}$\\
${}^{1}$Chinese Information Processing Laboratory ~ ${}^{2}$State Key Laboratory of Computer Science\\
Institute of Software, Chinese Academy of Sciences, Beijing, China\\
${}^{3}$University of Chinese Academy of Sciences, Beijing, China\\
${}^{4}$Baidu Inc., Beijing, China\\
\texttt{\{jialong2019,yaojie2017,hongyu,xianpei,sunle\}@iscas.ac.cn} \\
\texttt{\{xiaoxinyan,wu\_hua\}@baidu.com} \\
}

\date{}

\begin{document}
\maketitle
\begin{abstract}
{
  \renewcommand{\thefootnote}{\fnsymbol{footnote}}
}
One of the biggest bottlenecks in building accurate, high coverage neural open IE systems is the need for large labelled corpora. 
The diversity of open domain corpora and the variety of natural language expressions further exacerbate this problem.
In this paper, we propose a syntactic and semantic-driven learning approach, which can learn neural open IE models without any human-labelled data by leveraging syntactic and semantic knowledge as noisier, higher-level supervisions.
Specifically, we first employ syntactic patterns as data labelling functions and pretrain a base model using the generated labels. 
Then we propose a syntactic and semantic-driven reinforcement learning algorithm, which can effectively generalize the base model to open situations with high accuracy. 
Experimental results show that our approach significantly outperforms the supervised counterparts, and can even achieve competitive performance to supervised state-of-the-art (SoA) model.
\let\thefootnote\relax\footnotetext{${}^{*}$Corresponding author}
\end{abstract}

\section{Introduction} \label{sec:introduction}

\begin{figure}[t]
	\centering
	\includegraphics[scale=0.6]{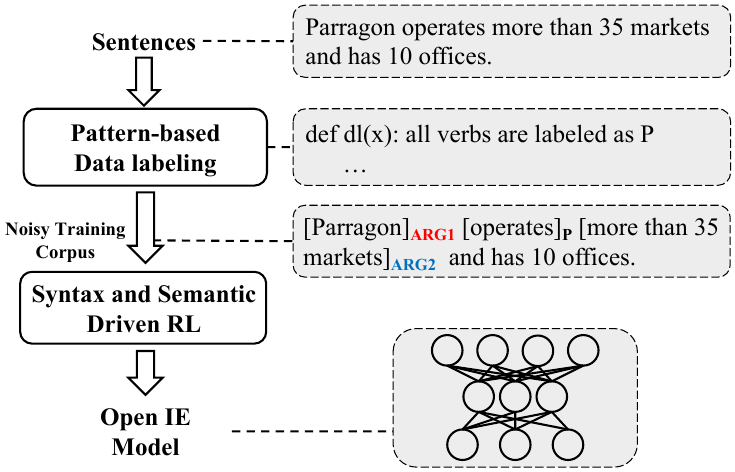}
  \caption{\label{fig:introduction}
    The proposed open IE framework, which consists of two learning strategies: 
    1) syntactic patterns are used as data labelling functions and a base model is pretrained using the generated labels; 
    2) a syntactic and semantic-driven RL algorithm is used to generalize the base model to open situations. 
	}
\end{figure}

Open information extraction (Open IE) aims to extract open-domain textual tuples consisting of a predicate and a set of arguments from massive and heterogeneous corpora~\citep{sekine-2006-demand,banko-etal-2007-open}.
For example, a system will extract a tuple (\emph{Parragon; operates; more than 35 markets}) from the sentence ``\emph{Parragon operates more than 35 markets and has 10 offices.}".
In contrary to the traditional IE, open IE is completely domain-independent and does not require the predetermined relations.

Recently, open IE has gained much attention~\citep{fader-etal-2011-identifying,akbik-loser-2012-kraken,mausam-etal-2012-open,corro-etal-2013-clausie,moro-2013-integrating,narasimhan-etal-2016-improving,pal-mausam-2016-demonyms,kadry-etal-2017-open,yu-etal-2017-open,roth-etal-2018-neural} and most of current open IE systems employ end-to-end neural networks, which first encode a sentence using Bi-LSTMs, then extract tuples by sequentially labelling all tokens in the sentence~\citep{stanovsky-etal-2018-supervised,jiang-etal-2019-improving,roy-etal-2019-supervising} or generating the target tuples token-by-token~\citep{zhang-etal-2017-mt,cui-etal-2018-neural,sun-etal-2018-logician}.
For example, to extract (\emph{Parragon; operates; more than 35 markets}), neural open IE systems will label the sentence as [\emph{B-$ARG_1$, B-P, B-$ARG_2$, I-$ARG_2$, I-$ARG_2$, I-$ARG_2$, O, O, O, O, O}] or generate a token sequence [\emph{$<$$ARG_1$$>$, Parragon, $<$$P$$>$, operates, $<$$ARG_2$$>$, more, than, 35, markets}]. 

The neural open IE systems, unfortunately, rely on the large labelled corpus to achieve good performance, which is often expensive and labour-intensive to obtain. 
Furthermore, open IE needs to extract relations of unlimited types from open domain corpus, which further exacerbates the need for large labelled corpus. 
Therefore, the labelled corpus is one of the biggest bottlenecks for neural open IE systems.

\begin{figure}[t]
  \centering
  \includegraphics[scale=0.675]{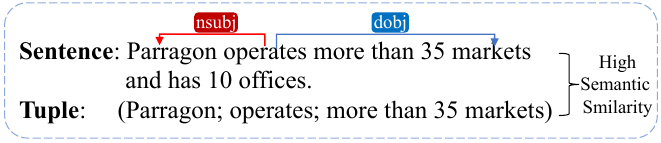}
  \caption{\label{fig:example}
  An example of open IE extractions, where the extracted tuple follows the \emph{nsubj} and \emph{dobj} dependency structure and is highly semantic similar to the original sentence. 
	}
\end{figure}

To resolve the labelled data bottleneck, this paper proposes a syntactic and semantic-driven learning approach, which can learn neural open IE models without any human-labelled data by leveraging syntactic and semantic knowledge as noisier, higher-level supervisions. 
The motivation of our method is that, although tuple extraction is a hard task, its inverse problem -- tuple assessment is easier to resolve by exploiting the syntactic regularities of relation expressions and the semantic consistency between a tuple and its original sentence. 
For example, Figure~\ref{fig:example} shows the ARG1 ``\emph{Parragon}" and the ARG2 ``\emph{more than 35 markets}" follow the \emph{nsubj} and \emph{dobj} dependency structure, respectively. 
Meanwhile, the extracted tuple (\emph{Parragon; operates; more than 35 markets}) has a high semantic similarity with its original sentence ``\emph{Parragon operates more than 35 markets and has 10 offices.}".
And we found that the syntactic regularities can be effectively captured using syntactic rules, and the semantic consistency can be effectively modelled using the recent powerful pre-trained models such as BERT~\citep{devlin-etal-2019-bert}.

Based on the above observations, we propose two learning  strategies to exploit syntactic and semantic knowledge for model learning. 
Figure~\ref{fig:introduction} illustrates the framework of our method. 
Firstly, syntactic open IE patterns are used as data labelling functions, and a base model is pretrained using the noisy training corpus generated by these labelling functions.
Secondly, because the pattern-based labels are often noisy and with limited coverage, we further propose a reinforcement learning algorithm which uses syntactic and semantic-driven reward functions, which can effectively generalize the base model to open situations with high accuracy. 
These two strategies together will ensure the effective learning of open IE models: the data labelling function can pretrain a reasonable initial model so that the RL algorithm can optimize model more effectively; although the pattern-based labels are often noisy and with low coverage, the RL algorithm can generalize the model to open situations with high accuracy. 

We conducted experiments on three open IE benchmarks: OIE2016~\citep{stanovsky-dagan-2016-creating}, WEB and NYT~\citep{mesquita-etal-2013-effectiveness}. 
Experimental results show that the proposed framework significantly outperforms the supervised counterparts, and can even achieve competitive performance with the supervised SoA approach. 
\footnote{Our source codes and experimental datasets are openly available at \url{https://github.com/TangJiaLong/SSD-OpenIE}.}

The main contributions of this paper are:
\begin{itemize}
\item We propose a syntactic and semantic-driven learning algorithm which can leverage syntactic and semantic knowledge as noisier, higher-level supervisions and learn neural open IE models without any human-labelled data.
\item We design two effective learning strategies for exploiting syntactic and semantic knowledge as supervisions: one is to use as data labelling functions and the other is to use as reward functions in RL. 
Experiments show that the two strategies are effective and can complement each other.
\item Because labelled data bottleneck is common in NLP tasks, we believe our syntactic and semantic-driven learning algorithm can motivate the learning of other NLP models, such as event extraction, etc.
\end{itemize}

\section{Syntactic and Semantic-driven Learning for Open IE}
In this section, we describe how to learn neural open IE models without any human-labelled data. 
Two strategies are proposed to exploit syntactic and semantic knowledge as noisier, higher-level supervisions. 
Firstly, the syntactic patterns are used as data labelling functions for heuristically labelling a training corpus. 
Secondly, the syntactic and semantic coherence scores between the extracted tuples and their original sentences are used as reward functions for reinforcement learning. 
These two strategies together will ensure the effective learning of open IE systems: 
1) although the labels generated by syntactic patterns are noisy and with limited coverage, they can pretrain a reasonable initial model; 
2) starting from the pretrained model, the syntactic and semantic-based reward functions provide an effective way to generalize our model to open situations.

In the following, we first introduce the neural networks used for open IE. 
Then we describe how to pretrain a base open IE model using syntactic patterns as data labelling functions. 
Finally, we generalize the base model using reinforcement learning with syntactic and semantic-driven rewards.

\subsection{Neural Open IE Model} \label{sec:neural}

This paper uses RnnOIE neural networks, which have shown its simplicity and effectiveness for open IE~\citep{stanovsky-etal-2018-supervised}.
But it should be noticed that our framework is not specialized to RnnOIE and can be used to train any neural open IE models.

RnnOIE formulates open IE as a sequence labelling task. 
Given a sentence $\bm{S} = (w_1 , w_2 , ..., w_m)$, RnnOIE will first identify all verbs in $\bm{S}$ as predicates, such as ``\emph{operates}" and ``\emph{has}" for ``\emph{Parragon operates more than 35 markets and has 10 offices.}". 
For each predicate $p$, RnnOIE will: 
1) first embed each word $w_i$ as $x_i = [e_i; \bm{I}(w_i = p)]$, where $e_i$ is $w_i$'s word embedding obtained by SoA pre-trained model BERT~\citep{devlin-etal-2019-bert}, and $\bm{I}(w_i = p)$ is an indicator vector which indicates whether $w_i$ is $p$; 
2) then obtain contextual word representations using a stacked BiLSTM with highway connections~\citep{srivastava-etal-2015-training,zhang-etal-2016-highway}: $\bm{H} = (h_1, h_2, ..., h_m) = BiLSTM(x_1, x_2, ..., x_m)$;
3) predict the probability of assigning label $y_i$ to a word $w_i$ using a fully connected feedforward classifier: $P(\hat{y}_i|\bm{S}, p, w_i) = softmax(\bm{W}h_i + \bm{b})$;
4) finally decode the full label sequence $\bm{\hat{Y}}$ using a beamsearch algorithm, \emph{e.g.}, RnnOIE will decode the label sequence [\emph{B-$ARG_1$, B-P, B-$ARG_2$, I-$ARG_2$, I-$ARG_2$, I-$ARG_2$, O, O, O, O, O}] to extract (\emph{Parragon; operates; more than 35 markets}). 

In open IE, all extracted tuples are ranked according to their confidence scores, which is important for downstream tasks, such as QA~\citep{fader-etal-2011-identifying} and KBP~\cite{angeli-etal-2015-leveraging}.
RnnOIE uses average log probabilities as the confidence of an extracted tuple:
\begin{equation} \label{fuc:avg}
  c(\bm{S}, p, \bm{\hat{Y}}) = \frac{\sum^{m}_{i=1}{log{P(\hat{y}_i|\bm{S}, p, w_i)}}}{m}
\end{equation}

Given a training corpus, RnnOIE can be supervisedly learned by maximum log-likelihood estimation (MLE):
\begin{equation} \label{fuc:object}
  \log{P(\bf{Y}|\bm{S}, p)} = \sum^{m}_{i=1}{\log{P(y_i|\bm{S}, p, w_i)}}
\end{equation} 
where $\bm{Y}=(y_1, y_2, ..., y_m)$ are the gold labels.
As discussed above, $\bm{Y}$ are expensive and labour-intensive to obtain and have become the biggest bottlenecks for neural open IE systems.
Therefore, it is critical to design a learning approach to get rid of this constraint. 

\subsection{Model Pretraining using Syntactic Pattern-based Data Labelling Functions} \label{sec:modelpretraining}

\begin{figure}[t]
  \centering
  \includegraphics[scale=0.65]{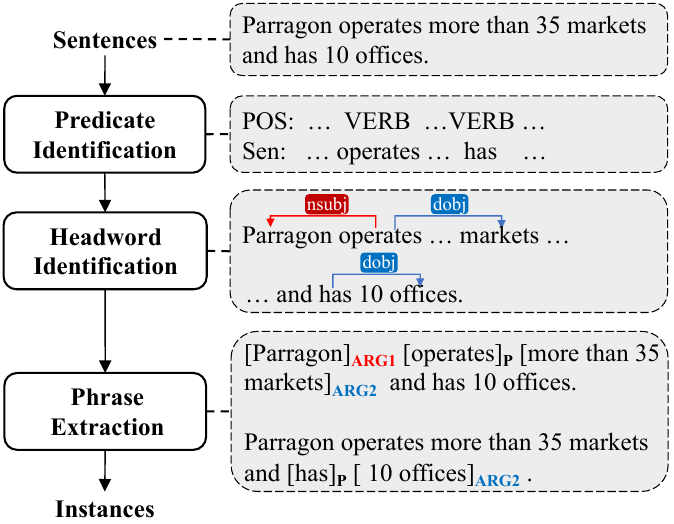}
  \caption{\label{fig:datalabelling}
    An overview of syntactic patterns as data labelling functions. 
    Two training instances are automatically generated using dependency pattern for predicates ``\emph{operates}" and ``\emph{has}". 
	}
\end{figure}

The first strategy is to use syntactic extraction patterns as data labelling functions, and then the heuristically labelled training corpus will be used to pretrain a neural open IE model.

It has long been observed that most relation tuples follow syntactic regularity, and many syntactic patterns have been designed for extracting tuples, such as TEXTRUNNER~\citep{banko-etal-2007-open} and ReVerb~\citep{fader-etal-2011-identifying}.
However, it is difficult to design high coverage syntactic patterns, although many extensions have been proposed, such as WOE~\citep{wu-weld-2010-open}, OLLIE~\citep{mausam-etal-2012-open}, ClausIE~\citep{corro-etal-2013-clausie}, Standford Open IE~\citep{angeli-etal-2015-leveraging}, PropS ~\citep{stanovsky-etal-2016-getting} and OpenIE4~\citep{mausam-2016-open}.

This paper leverages the power of patterns differently. 
Inspired by the ideas of data programming~\citep{ratner-etal-2016-data} and distant supervision~\citep{mintz-etal-2009-distant}, we use syntactic patterns as data labelling functions, rather than to directly extracting tuples.

Concretely, this paper uses dependency patterns from Standford Open IE~\citep{angeli-etal-2015-leveraging} to design hand-crafted patterns as data labelling functions.
As shown in Figure~\ref{fig:datalabelling}, given a sentence and its dependency parse, two training instances are generated:
1) We first identify all its predicates using part of speech (POS) tags. For example,  ``\emph{operates}" and ``\emph{has}" are identified. 
2) For each predicate, we identify its arguments' headwords using predefined dependency patterns \footnote{The dependency relations are defined in \url{https://nlp.stanford.edu/software/dependencies_manual.pdf}}. 
For example, ``\emph{Parragon}" and ``\emph{markets}" are extracted as the headwords. 
3) For each headword, we extract the whole phrase headed to it as subject/object. 
For example, the phrase ``\emph{more than 35 markets}" headed to ``\emph{markets}" will be extracted as the object of ``\emph{operates}". 

Finally, the generated labels are used to pretrain an open IE model by optimizing the objective function (\ref{fuc:object}), which can provide a reasonable initialization for starting our RL algorithm in the next section.

\subsection{Model Generalization via Syntactic and Semantic-driven Reinforcement Learning} \label{sec:model_generalization}

One main drawback of the automatically generated labels is that they are often noisy and with limited coverage, \emph{i.e.}, many open relation tuples are not covered by the predefined patterns, and the dependency parse may contain errors which in turn will lead to noisy training instances. 
For example, in Figure~\ref{fig:datalabelling} the training instance of the predicate ``\emph{has}" misses its subject ``\emph{Parragon}".  
Therefore, it is critical to generalize and refine the base model to open situations for good performance.

To this end, this section proposes the second learning strategy: syntactic and semantic-driven reinforcement learning. 
Specifically, we first measure the goodness of extracted tuples based on syntactic constraints using syntactic rules and semantic consistencies using pre-trained models such as BERT~\citep{devlin-etal-2019-bert}.
And then we generalize our model using the goodness of extractions as rewards in RL. 

By modelling the extraction task as a Markov Decision Process (MDP), we have the following definitions: $<\bm{S}, \bm{A}, \bm{T}, \bm{R}>$:
\begin{itemize}
\item $\bm{S} = \{\bm{s}\}$ are states used to capture the information from the current sentence. Specifically, $\bm{S}$ are hidden states $\bm{H}$ obtained by stacked BiLSTM.
\item $\bm{A} = \{\bm{a}\}$ are actions used to indicate the target labels which are decided based on the current states $\bm{S}$ and the beam search strategy.
\item $\bm{T}$ is the state transition function, which is related to the state update.
\item $\bm{R}(\bm{\hat{Y}}, \bm{S})$ is the reward function, which models the goodness of the extracted tuples. We will detailly describe our syntactic and semantic-driven reward function in the next paragraph.
\end{itemize}
Formally, the open IE model is trained to maximize the expected reward of the generated label sequence $\bf{\hat{Y}}$ using the REINFORCE algorithm with likelihood ratio trick~\citep{glynn-1990-likelihood,williams-1992-simple}:
\begin{align} \nonumber
  \nabla J(\theta) &= E_{\bm{\hat{Y}} \backsim P(\bm{\hat{Y}}|\bm{S},\bm{p})} [\bm{R}(\bm{\hat{Y}}, \bm{S})] \\
  & \approx \bm{R}(\bm{\hat{Y}}, \bm{S}) \nabla \log{P(\bf{\hat{Y}}|\bm{S}, p)} 
\end{align}
where $\log{P(\bf{\hat{Y}}|\bm{S}, p)}$ denotes the probability of the generated label sequence. 

\paragraph{Reward Function.} 
The reward function, i.e., the goodness of extracted tuples, is critical in our RL algorithm. 
This paper estimates the reward $\bm{R}(\bm{\hat{Y}}, \bm{S})$ by considering both syntactic constraint and semantic consistency:
\begin{equation} \label{fuc:r}
  \bm{R}(\bm{\hat{Y}}, \bm{S}) = Syn(\bm{\hat{Y}}) * Sem(\bm{\hat{Y}}, \bm{S})
\end{equation} 
where $Syn(\bm{\hat{Y}})$ is the syntactic constraint score  and $Sem(\bm{\hat{Y}}, \bm{S})$ is the semantic consistency score.

Following~\citet{he-etal-2015-question,stanovsky-etal-2018-supervised,jiang-etal-2019-improving}, we judge an extracted tuple as correct if and only if it's predicate and arguments include their corresponding syntactic headwords (Headwords Match).
Otherwise, the extracted tuples are judged as incorrect.
That is:
\begin{align} \label{fuc:syn}
  & Syn(\bm{\hat{Y}}) = \begin{cases}
      1, & \mbox{Headwords Match}\\
      -1, & \mbox{Else}
      \end{cases}
\end{align}
where 1 means the predicted label sequence $\hat{\bm{Y}}$ is correct and -1 for incorrect. 

For semantic consistency, given an extracted relation and its original sentence, $Sem(\bm{\hat{Y}}, \bm{S})$ is computed as:
\begin{equation}\label{fuc:sem}
  Sem(\bm{\hat{Y}}, \bm{S}) = P(positive|\bm{\hat{Y}}, \bm{S}) 
\end{equation}
where $P(positive|\bm{\hat{Y}}, \bm{S})$ is the semantic similarity between the predicted label sequence $\hat{\bm{Y}}$ and its original sentence $\bm{S}$.
This paper estimates this semantic similarity using a BERT-based classifier, which assigns a similarity score to each sentence-tuple pair.
Because multiple tuples can be extracted from a single sentence (see Figure~\ref{fig:datalabelling} for example), we train the 
classifier using the Stanford Natural Language Inference (SNLI) Corpus~\citep{bowman-etal-2015-large}, so that a high similarity score will be assigned if the original sentence entails the extracted tuple. 
This semantic consistency can provide useful supervision signals for open IE models. 
For example, because (\emph{Parragon; has; 10 offices}) has higher semantic similarity than (\emph{has; 10 offices}) to sentence ``\emph{Parragon operates more than 35 markets and has 10 offices.}", the model will be guided to more complete extractions.

\paragraph{Semantic-Based Confidence Estimation.} 
In RnnOIE, the confidence score $c(\bm{S}, p, \bm{\hat{Y}})$ is estimated only using extraction probabilities. 
This paper further considers the semantic consistency score for better confidence estimation:
\begin{equation} \label{fuc:confidence}
  c'(\bm{S}, p, \bm{\hat{Y}}) = c(\bm{S}, p, \bm{\hat{Y}}) + log(Sem(\bm{\hat{Y}}, \bm{S}))
\end{equation}
where the $log$ is used for semantic consistency because $c(\bm{S}, p, \bm{\hat{Y}})$ also uses $log$ probabilities.

\section{Experiments}

\subsection{Experimental Settings}

\paragraph{Datasets.}
We conduct experiments on three open IE benchmarks: OIE2016~\citep{stanovsky-dagan-2016-creating}, WEB and NYT~\citep{mesquita-etal-2013-effectiveness}. 
Table~\ref{tab:datasets} shows their statistics. 
Because only OIE2016 provides training instances and it is the largest dataset, we use OIE2016 as the primary dataset. 
The WEB and NYT datasets are small and without training instances, therefore we use them for out-of-domain evaluation. 
For OIE2016, we follow the settings in~\citet{jiang-etal-2019-improving}. 
For WEB and NYT, we follow the settings in~\citet{stanovsky-etal-2018-supervised}.

\begin{table}[t]
  \centering
  \resizebox{0.49\textwidth}{!}{
  \begin{tabular}{c|c|c|c|c}
  \hline
  \textbf{Dataset} & \textbf{Type} & \textbf{Train} & \textbf{Dev} & \textbf{Test}\\
  \hline
  \hline
  \multirow{2}*{OIE2016}  & sentence   & \multicolumn{1}{r|}{1,688}  & \multicolumn{1}{r|}{560} & \multicolumn{1}{r}{641} \\
                          & relation  & \multicolumn{1}{r|}{3,040}  & \multicolumn{1}{r|}{971} & \multicolumn{1}{r}{1,729} \\
  \hline
  \multirow{2}*{WEB}      & sentence   & -- & -- & \multicolumn{1}{r}{500}\\
                          & relation  & -- & -- & \multicolumn{1}{r}{461} \\
  \hline
  \multirow{2}*{NYT}      & sentence   & -- & -- & \multicolumn{1}{r}{222}\\
                          & relation  & -- & -- & \multicolumn{1}{r}{222} \\
  \hline
  \end{tabular}}
  \caption{\label{tab:datasets}
  Statistics of OIE2016, WEB and NYT.
  }
\end{table}

\subsection{Baselines}

We compare our method with the following baselines:
\begin{itemize}
  \item \textbf{Pattern-based open IE systems} which utilize syntactic patterns to extract relations, including \emph{ClausIE}~\citep{corro-etal-2013-clausie}, \emph{StandfordOpenIE}~\citep{angeli-etal-2015-leveraging}, \emph{PropS}~\citep{stanovsky-etal-2016-getting} and \emph{OpenIE4}~\citep{mausam-2016-open}. 
  \item \textbf{Supervised neural open IE systems}, including \emph{RnnOIE-Supervised}~\citep{stanovsky-etal-2018-supervised} and \emph{RankAware}~\citep{jiang-etal-2019-improving}.
  \emph{RnnOIE} is described in Section~\ref{sec:neural}.
  \emph{RankAware} is the state-of-the-art model in OIE2016 dataset, which uses iterative rank-aware learning for better confidence estimation.
\end{itemize}

\subsection{Overall Results}

\begin{figure}[t]
  \centering
  \includegraphics[scale=0.68]{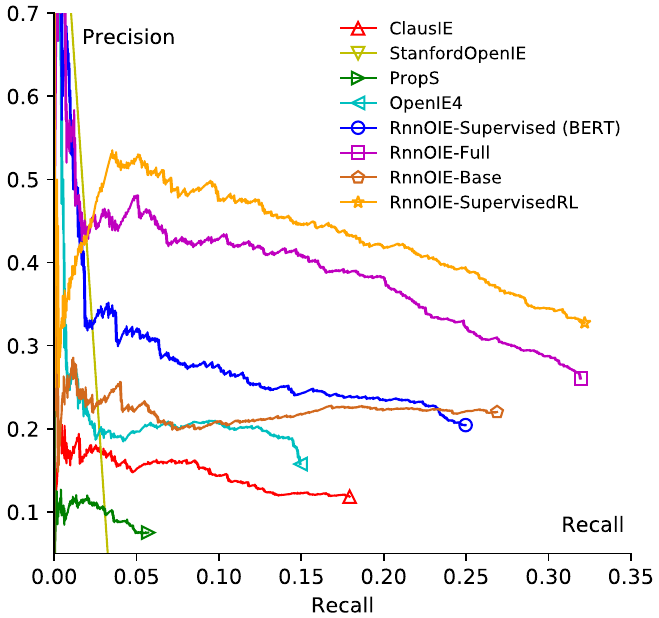}
  \caption{
  \label{fig:oie2016_pr}
  PR curves of different systems on OIE2016.
  }
\end{figure}

Table~\ref{tab:overall} and Figure~\ref{fig:oie2016_pr} shows the overall results. 
For our method, we use three settings: 
the first is the full model using the proposed syntactic and semantic-driven learning -- \emph{RnnOIE-Full}; 
the second is the base model which is not generalized using our reinforcement learning strategy -- \emph{RnnOIE-Base}; 
the third is our method with the base model trained using a gold-labelled corpus -- \emph{RnnOIE-SupervisedRL}. 
From Table~\ref{tab:overall} and Figure~\ref{fig:oie2016_pr}, we can see that:
\footnote{The performance of \emph{RnnOIE-Supervised} reported in our paper is lower than the original paper~\citep{stanovsky-etal-2018-supervised} because the authors use a more lenient lexical overlap metric in their released code: https://github.com/gabrielStanovsky/oie-benchmark. Following \citet{jiang-etal-2019-improving}, we judge an extraction as correct if the predicate and arguments include the syntactic head of the gold standard counterparts.}

\begin{table*}[t]
  \renewcommand\baselinestretch{1.1}
  \centering
  \resizebox{0.99\textwidth}{!}{
  \begin{tabular}{ l|r|r|r|r|r|r }
  \hline
  \multirow{2}*{\textbf{Model}} & \multicolumn{2}{c|}{\textbf{OIE2016}} & \multicolumn{2}{c|}{\textbf{WEB}} & \multicolumn{2}{c}{\textbf{NYT}} \\
  \cline{2-7}
   & \multicolumn{1}{c|}{AUC} & \multicolumn{1}{c|}{F1} & \multicolumn{1}{c|}{AUC} & \multicolumn{1}{c|}{F1} & \multicolumn{1}{c|}{AUC} & \multicolumn{1}{c}{F1} \\
  \hline
  \hline
  \textbf{Pattern-based}  &  &  &  &  &  & \\
  \quad ClausIE~\citep{corro-etal-2013-clausie} &  2.6 & 14.4 & 11.3 & 24.3 &  1.5 &  6.1 \\
  \quad StandfordOpenIE~\citep{angeli-etal-2015-leveraging} &  1.7 & 3.2 & --- & --- & --- & --- \\
  \quad PropS~\citep{stanovsky-etal-2016-getting}  &  0.6 &  6.5 &  0.3 &  4.8 &  0.2 &  3.3 \\
  \quad OpenIE4~\citep{mausam-2016-open} &  3.4 & 16.4 &  7.1 & 27.7 &  2.3 & \textbf{15.8} \\
  \hline
  \hline
  \textbf{Supervised Learning}  &  &  &  &  &  & \\
  \quad RnnOIE-Supervised (original)~\citep{stanovsky-etal-2018-supervised} &  5.0 & 20.4 &  --- &  --- &  --- &  --- \\
  \quad RnnOIE-Supervised (BERT) &  7.2 & 22.9 &  3.3 & 16.0 &  0.9 &  8.4 \\
  \quad RankAware~\citep{jiang-etal-2019-improving} & 12.5 & 31.5 &  --- &  --- &  --- &  --- \\
  \hline
  \hline
  \textbf{Syntactic and Semantic-driven Learning}  &  &  &  &  &  & \\
  \quad RnnOIE-Full (Pretrained base model with RL)  & 13.8 & \textbf{32.5} & \textbf{15.8} & \textbf{37.9} & \textbf{2.6}  & 14.3  \\
  \quad RnnOIE-Base (Pretrained base model w/o RL)  &  5.9 & 24.2 & 10.5 & 31.7 &  1.8 & 10.3 \\
  \quad RnnOIE-SupervisedRL (Supervised base model with RL)  & \textbf{15.9}  & \textbf{32.2} & 11.2  & 29.4 & \textbf{2.6} & 11.4 \\
  \hline
  \end{tabular}}
  \caption{\label{tab:overall}
  The overall results on OIE2016, WEB and NYT. 
  For fair comparison, all results of baselines are adapted from their original papers except the BERT version \emph{RnnOIE} -- \emph{RnnOIE-BERT}.
  }
\end{table*}

\textbf{1) The syntactic and semantic-driven learning approach can effectively resolve the training data bottleneck of neural open IE systems.} 
In all three datasets, \emph{RnnOIE-Full} significantly outperforms its supervised counterpart -- \emph{RnnOIE-Supervised (BERT)}. 
On OIE2016, \emph{RnnOIE-Full} can even achieve competitive performance with the supervised SoA model -- \emph{RankAware}. 
We believe this verifies the motivation of our method: 
the quality of extractions can be accurately evaluated using syntactic and semantic knowledge, and this knowledge can be effectively leveraged for the learning of open IE systems.

\textbf{2) Syntactic pattern-based data labelling is an effective learning strategy.}
By generating training corpus, \emph{RnnOIE-Base} achieves competitive performance on OIE2016 compared with its supervised counterpart -- \emph{RnnOIE-Supervised (BERT)}
This verifies that the heuristically labelled dataset, although may noisy, can also provide a good start for building open IE systems. 
On the other side, we found noisy training corpus itself is not enough for high-performance open IE systems: in OIE2016 there is a 134\% AUC gap (5.9 to 13.8) from \emph{RnnOIE-Base} to \emph{RnnOIE-Full}. 
This also verifies the need for further generalization techniques.

\textbf{3) Syntactic and Semantic-driven RL is effective for generalize and refine open IE models.} 
Compared with \emph{RnnOIE-Base}, \emph{RnnOIE-Full} can get a 134\% AUC improvement, from 5.9 to 13.8. 
By further generalizing the supervised \emph{RnnOIE-Supervised (BERT)} baseline using RL, \emph{RnnOIE-SupervisedRL} can further obtain a 121\% AUC improvement, from 7.2 to 15.9. 
The above results verify the effectiveness of our RL algorithm, and this may be because
a) the RL is based on the explore-and-exploit strategy, and the explore stage can consider many unseen cases; 
b) the syntactic and semantic knowledge is good supervision signals for open IE systems, and the syntactic and semantic-aware rewards can effectively exploit these signals.

\textbf{4) The RL-based generalization strategy is critical for scaling open IE systems to open situations.}
In OIE2016, we can see that, although supervised systems can outperform pattern-based systems, their performance decreases significantly in out-of-domain WEB and NYT datasets. 
\emph{RnnOIE-Supervised (BERT)} even perform worse than \emph{ClausIE} and \emph{OpenIE4} on WEB and NYT. 
On the contrary, \emph{RnnOIE-Full} can still achieve robust performance. 
This verifies the effectiveness of the proposed RL-based algorithm for generalizing to open situations.
It is worth to notice that \emph{RnnOIE-Full} even outperforms \emph{RnnOIE-SupervisedRL} on out-of-domain datasets.
The reason behind it may be: 
a)	The gold-labelled corpus is useful in in-domain situations (OIE2016). However, supervised base model may be overfitting and further affects the generalization process in RL. 
b) \emph{RnnOIE-Full} learns shallow linguistic features which are more general. Therefore it performs better in out-of-domain situations (WEB and NYT).

\subsection{Detailed Analysis}
To analyze our method in detail, this section further investigates the effects of syntactic and semantic knowledge, semantic-based confidence estimation and RL exploration beam size.

Additionally, we compare \emph{RnnOIE-Full} with two open IE systems to find out how far can data labelling functions get us.

\begin{table}[t]
  \renewcommand\baselinestretch{1.1}
  \centering
  \resizebox{0.475\textwidth}{!}{
  \begin{tabular}{l|r|r|r|r}
  \hline
   & \multicolumn{1}{c|}{\textbf{AUC}} & \multicolumn{1}{c|}{$\bf \Delta_{AUC}$} & \multicolumn{1}{c|}{\textbf{F1}} & \multicolumn{1}{c}{$\bf \Delta_{F1}$} \\
  \hline
  RnnOIE-Full              &  13.8 &   & 32.5 &  \\
  \quad w/o semantic      &  12.3 & -10.9\% & 31.7 & -2.5\% \\
  \quad w/o syntactic     &  3.0 & -78.3\% & 16.9 & -48.0\% \\
  \hline
  \end{tabular}
  }
  \caption{\label{tab:ablation}
  The performance of \emph{RnnOIE-Full} with different reward settings on OIE2016.
  }
\end{table}

\paragraph{Effects of Syntactic and Semantic Knowledge.}
Our reward function $\bm{R}(\bm{\hat{Y}}, \bm{S})$ consists of both syntactic constraint $Syn(\bm{\hat{Y}})$ and semantic consistency $Sem(\bm{\hat{Y}}, \bm{S})$.
To analyze the effects of syntactic and semantic knowledge, we conduct ablation experiments by removing the semantic part (\emph{w/o semantic}) and removing the syntactic part (\emph{w/o syntactic}) in reward function. 
Table~\ref{tab:ablation} shows their performances on OIE2016.

We can see that: 
1) both syntactic and semantic knowledge are useful: removing any of them will result in a performance decrease; 
2) syntactic constraint is crucial for our model: removing it will result in a significant AUC decrease (from 13.8 to 3.0). 
This is because if we drop syntactic constraint $Syn(\bm{\hat{Y}})$, all explored relations will be treated as true, therefore our RL algorithm cannot rectify the wrong extractions.

\begin{table}[t]
  \renewcommand\baselinestretch{1.1}
  \centering
  \resizebox{0.475\textwidth}{!}{
  \begin{tabular}{l|r|r}
  \hline
  \textbf{Confidence Estimation Algorithm} & \multicolumn{1}{c|}{\textbf{AUC}} & \multicolumn{1}{c}{\textbf{F1}}  \\
  \hline
  Avg Log & 12.0  & 29.1  \\
  Semanctic Consistency & 10.8 & \bf{32.5}  \\
  Avg Log + Semanctic Consistency & \bf{13.8} & \bf{32.5}  \\
  \hline
  \end{tabular}
  }
  \caption{\label{tab:confidence}
  The performance of \emph{RnnOIE-Full} with different confidence estimation settings on OIE2016.
  }
\end{table}

\paragraph{Effect of Confidence Estimation.}
Table~\ref{tab:confidence} shows the performance using different confidence estimation algorithms, including:
\emph{Avg Log} (average log probabilities) which is computed as Function~\ref{fuc:avg}, \emph{Semantic Consistency} which is computed as Function~\ref{fuc:sem} and \emph{Avg Log + Semantic Consistency}.

We can see that: 
1) the semantic evidence and the model prediction evidence are complementary to each other: \emph{Avg Log + Semantic Consistency} obtains the best performance and gets 15\% and 28\% AUC improvements to \emph{Avg Log} and \emph{Semantic Consistency}; 
2) \emph{Semantic Consistency} can provide useful information for confidence estimation: \emph{Semantic Consistency} itself can achieve comparable performance with \emph{Avg Log}.

\begin{figure}[t]
  \centering
  \includegraphics[scale=0.8]{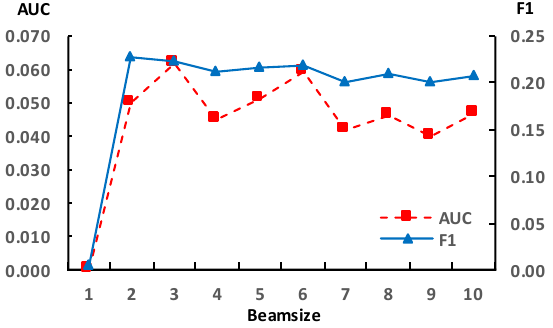}
  \caption{
  \label{fig:beamsize}
  AUC and F1-scores of \emph{RnnOIE-Full} with different beam sizes on the OIE2016 validation set.
  }
\end{figure}

\paragraph{Effect of Beam Size.}
The beam size is an important hyper-parameter which controls the exploration breadth of our RL algorithm. 
Figure~\ref{fig:beamsize} shows the performance with different beam sizes.

We can see that: 
1) an appropriate beam size is needed for generalizing open IE model. 
If the beam size is too small, \emph{RnnOIE-Full} cannot explore new unseen cases because its explore strategy is too greedy; 
late2) The proposed RL algorithm is robust and achieves good performance with reasonable beam sizes ($\geq$ 3).
Because larger beam size will increase the computational complexity, we set beam size to 3 in all other experiments.

\begin{table}[t]
  \renewcommand\baselinestretch{1.1}
  \centering
  \resizebox{0.475\textwidth}{!}{
  \begin{tabular}{l|r|r}
  \hline
  \textbf{Model} & \multicolumn{1}{c|}{\textbf{AUC}} & \multicolumn{1}{c}{\textbf{F1}}  \\
  \hline
  NeuralOpenIE~\cite{cui-etal-2018-neural} & 47.3 & ---  \\
  SencseOIE~\cite{roy-etal-2019-supervising} & --- & 70.0  \\
  RnnOIE-Full & \bf{56.0} & \bf{76.7}  \\
  \hline
  \end{tabular}
  }
  \caption{\label{tab:complex}
  The results evaluated by Lexical Overlap on OIE2016. 
  For fair comparison, all results of baselines are adapted from their original papers.
  }
\end{table}

\begin{table*}[t]
  \renewcommand\baselinestretch{1.1}
  \centering
  \resizebox{0.99\textwidth}{!}{
  \begin{tabular}{l|l}
  \hline
  \textbf{Error} & \textbf{Sentence} \\
  \hline
  \hline
  \bf{Missing Argument} & [DePauw University]\textcolor[rgb]{0,0,1}{\scriptsize{ARG1}} [awarded]\textcolor[rgb]{0,0,1}{\scriptsize{P}} [the degree `` Doctor of Divinity]\textcolor[rgb]{0,0,1}{\scriptsize{ARG2}} '' [\colorbox[rgb]{1,0.25,0.25}{in 1892}]\textcolor[rgb]{1,0.25,0.25}{\scriptsize{Missing}} .  \\
  \hline
  \hline
  \bf{Overgenerated Predicate} & [A British version of this show was developed , known as `` Gladiators : Train 2]\textcolor[rgb]{0,0,1}{\scriptsize{ARG1}} [\colorbox[rgb]{1,0.25,0.25}{Win}]\textcolor[rgb]{1,0.25,0.25}{\scriptsize{Overgenerated}} '' . \\
  \hline
  \hline
  \bf{Incorrect Annotation} & Coke has tended to increase its control [when results were sluggish]\textcolor[rgb]{0,0,1}{\scriptsize{ARG2}} in a [\colorbox[rgb]{1,0.25,0.25}{given}]\textcolor[rgb]{1,0,0}{\scriptsize{Incorrect annotated}} [country]\textcolor[rgb]{0,0,1}{\scriptsize{ARG1}} . \\
  \hline
  \end{tabular}
  }
  \caption{\label{tab:error}
  Bad cases of the proposed model \emph{RnnOIE-Full}.
  }
\end{table*}

\paragraph{More Complex Data Labelling Functions.}
In section~\ref{sec:modelpretraining}, we directly use dependency patterns from Standford Open IE~\citep{angeli-etal-2015-leveraging} to design hand-crafted patterns as data labelling functions.
It raises a question that, if we use more complex patterns as data labelling functions to obtain more diverse and accurate labelled data, is it still necessary to move on to RL approach?
To answer this question, we compare \emph{RnnOIE-Full} with two open IE systems, NeuralOpenIE~\cite{cui-etal-2018-neural} and SencseOIE~\cite{roy-etal-2019-supervising}, to find out how far can data labelling functions get us.

From Table~\ref{tab:complex}, we can see that: \citet{cui-etal-2018-neural} formulates open IE as a sentence generation task and uses OpenIE4~\cite{mausam-2016-open} to generate train examples (AUC 47.3); 
\citet{roy-etal-2019-supervising} uses three open IE systems to extract additional features to enrich human labelled train examples (F1 70.0 without other defined embedding features). Different from them, \emph{RnnOIE-Full} does not use any labelled data and includes model generalization via RL (AUC 56.0; F1 76.7).
This verifies the effectiveness and the necessity of the proposed RL-based algorithm.

\subsection{Error Analysis.}

We further conduct error analysis for \emph{RnnOIE-Full}. 
We found there are mainly three types of error cases: \emph{Missing Argument}, \emph{Overgenerated Predicate} and \emph{Wrong Annotation}. 
Table~\ref{tab:error} shows their examples. 

\emph{Missing Argument} is the case where the extractions miss some arguments, especially for some optional arguments such as \emph{Time} and \emph{Place} in \emph{RnnOIE-Full}. 
For instance, the first case in Table~\ref{tab:error} shows the extraction for predicate ``\emph{award}" misses the optional time argument ``\emph{in 1892}", although it correctly contains two main arguments ``\emph{DePauw University}" and ``\emph{the degree ``Doctor of Divinity"}". 
We found this maybe because optional arguments usually play a less important role in semantic consistency, our syntactic and semantic-driven RL algorithm will pay less attention to this generalization.

\emph{Overgenerated Predicate} is the case where the predicates of extractions are not included in the ground truth.
The second case in Table~\ref{tab:error} shows a bad case where ``\emph{Win}" is wrongly extracted as the predicate. 
This is a common error in all neural-based approaches because they generally treat all verbs in a sentence as predicates and do not have a mechanism to reject incorrect ones.
One strategy to handle this error is to jointly detect predicates and arguments, which we leave as future work.  

\emph{Incorrect Annotation} is the case where the ground truth labels are incorrect. 
Because expressions in open IE are highly diversified, we found the gold annotations may be incorrect or inconsistent. 
The third case in Table~\ref{tab:error} shows an incorrect ground truth annotation ``\emph{given}", which is wrongly labelled as a predicate.  
This further verifies the bottleneck of high-quality, large scale labelled corpus for open IE. 

\section{Related Work}
\paragraph{Open IE.}
Open IE approaches can be mainly categorized into two categories: pattern-based and neural-based. 
Pattern-based open IE approaches extract relational tuples using syntactic patterns~\citep{banko-etal-2007-open,fader-etal-2011-identifying,wu-weld-2010-open,mausam-etal-2012-open,mausam-2016-open,corro-etal-2013-clausie,angeli-etal-2015-leveraging,stanovsky-etal-2016-getting}; 
In recent years, neural-based approaches have achieved significant progress, which formulate open IE as either a sequence labelling task~\citet{stanovsky-etal-2018-supervised,jiang-etal-2019-improving,roy-etal-2019-supervising} or a sentence generation task via encoder-decoder framework~\citet{cui-etal-2018-neural,zhang-etal-2017-mt,sun-etal-2018-logician}.



Syntactic and semantic knowledge has also been leveraged to enhance open IE systems.  
\citet{moro-2013-integrating} design additional syntactic and semantic features to enhance their kernel-based open IE system.
\citet{roy-etal-2019-supervising} incorporate the outputs of multiple pattern-based Open IE systems as additional features to supervised neural open IE systems to overcome the problem of insufficient.
Compared with these studies which exploit syntactic and semantic knowledge as additional features of a supervised system, this paper exploits syntactic and semantic knowledge as supervision signals, so that neural open IE models can be effectively learned without any labelled data.

\paragraph{Data Augmentation for NLP.}
The labelled data bottleneck is a common problem in NLP, therefore many data augmentation techniques have been proposed, such as data programming~\citep{ratner-etal-2016-data}, distant supervision~\citep{mintz-etal-2009-distant}.  
Data programming paradigm~\citep{ratner-etal-2016-data} creates training datasets by explicitly representing users' expressions or domain heuristics as a generative model. 
Distant supervision paradigm~\citep{mintz-etal-2009-distant} heuristically generates labelled dataset by aligning facts in KB with sentences in the corpus. 
The proposed data labelling functions are also motivated by the ideas of data programming and distant supervision.

\paragraph{Reinforcement Learning for IE.}
Reinforcement learning (RL)~\citep{sutton-barto-1998-reinforcement} follows the explore and exploit paradigm and is apt for optimizing non-derivative learning objectives in NLP~\citep{wu-etal-2018-study}.
Recently, RL has gained much attention in information extraction~\citep{qin-etal-2018-robust,qin-etal-2018-dsgan,takanobu-2019-hierarchical}. In open IE, \citet{narasimhan-etal-2016-improving} firstly using traditional Q-learning method to extract textual tuples.
However, their reward function is chosen to maximize the final extraction accuracy which still relies on human-labelled datasets and can not capture the syntactic and semantic supervisions explicitly.

\section{Conclusions}
This paper proposes an open IE learning approach, which can learn neural models without any human-labelled data by leveraging syntactic and semantic knowledge as noisier, higher-level supervisions. 
Specifically, two effective learning strategies are proposed, including the pattern-based data labelling functions and the syntactic and semantic-driven RL algorithm. 
Experimental results show that our method significantly outperforms supervised counterparts, and can even achieve competitive performance to supervised SoA model. 
Furthermore, because labelled data is a common bottleneck in NLP, we believe our syntactic and semantic-driven learning approach can also be used for other NLP tasks, such as event extraction, etc.

\section{Acknowledgments}
This research work is supported by the National Natural Science Foundation of China under Grants no. U1936207, National Key R\&D Program of China under Grant 2018YFB1005100, the National Key Research and Development Project of China (No. 2018AAA0101900), Beijing Academy of Artiﬁcial Intelligence (BAAI2019QN0502), and in part by the Youth Innovation Promotion Association CAS(2018141).

\bibliography{anthology,emnlp2020}

\end{document}